%% file: iclr2023_conference.tex
\documentclass{article} % For LaTeX2e
\usepackage{iclr2023_conference,times}
\usepackage{graphicx}
\usepackage{booktabs}
\input{math_commands.tex}

\usepackage{spverbatim}
\usepackage[framemethod=TikZ]{mdframed}
\usepackage{hyperref}
\usepackage{url}

\iclrfinaltrue

\title{Natural Language Reinforcement Learning}

\author{%
\textbf{Xidong Feng$^{1}$\thanks{{{Equal Contribution. Correspondence to xidong.feng.20@ucl.ac.uk.}}},  Ziyu Wan$^{2*}$, Mengyue Yang$^{1}$, Ziyan Wang$^{3}$, Girish A. Koushik$^{4}$, } \\ \textbf{Yali Du$^3$, Ying Wen$^2$, Jun Wang$^1$} \\ 
  $^1$University College London, $^2$Shanghai Jiao Tong University,  $^3$King's Collge London, \\ $^4$University of Surrey}

\begin{document}

\maketitle

\begin{abstract}
Reinforcement Learning (RL) has shown remarkable abilities in learning policies for decision-making tasks. However, RL is often hindered by issues such as low sample efficiency, lack of interpretability, and sparse supervision signals. To tackle these limitations, we take inspiration from the human learning process and introduce \textbf{Natural Language Reinforcement Learning (NLRL)}, which innovatively combines RL principles with natural language representation. Specifically, NLRL redefines RL concepts like task objectives, policy, value function, Bellman equation, and policy iteration in natural language space. We present how NLRL can be practically implemented with the latest advancements in large language models (LLMs) like GPT-4. Initial experiments over tabular MDPs demonstrate the effectiveness, efficiency, and also interpretability of the NLRL framework.
\end{abstract}

\section{Introduction}
Reinforcement Learning (RL) constructs a mathematical framework that encapsulates key decision-making elements. It quantifies the objectives of tasks through the concept of cumulative rewards, formulates policies with probability distributions, expresses value functions via mathematical expectations, and models environment dynamics through state transition and reward functions. This framework effectively converts the policy learning problem into an optimization problem.

Despite the remarkable achievements of RL in recent years, significant challenges still underscore the framework's limitations. For example, RL suffers from the sample efficiency problem--RL algorithms are task-agnostic and do not leverage any prior knowledge, requiring large-scale and extensive sampling to develop an understanding of the environment. RL also lacks interpretability. Despite the superhuman performance of models like AlphaZero \citep{silver2017alphazero} in mastering complex games such as Go, the underlying strategic logic of their decision-making processes remains elusive, even to professional players. In addition, the supervision signal of RL is a one-dimensional scalar value, which is much more sparse compared with traditional supervised learning over information-rich datasets such as texts and images. This is also one of the reasons for the instability of RL training \citep{zheng2023secrets, andrychowicz2020matters}.

These limitations drive us to a new framework inspired by the human learning process. Instead of mathematically modeling decision-making components like RL algorithms, humans tend to conduct relatively vague operations by natural language. First, natural language enables humans with text-based prior knowledge, which largely increases the sample efficiency when learning new tasks. 
Second, humans possess the unique ability to articulate their explicit strategic reasoning and thoughts in natural language before deciding on their actions, making their process fully interpretable by others, even if it's not always the most effective approach for task completion. Third, natural language data contains information about thinking, analysis, evaluation, and future planning. It can provide signals with high information density, far surpassing that found in the reward signals of traditional RL.
Inspired by the human learning process, we propose \textbf{Natural Language Reinforcement Learning} (NLRL), a new RL paradigm that innovatively combines the traditional RL concepts and natural language representation. By transforming key RL components—such as task objectives, policies, value functions, the Bellman equation, and generalized policy iteration (GPI) \citep{sutton2018reinforcement}—into their natural language equivalents, we harness the intuitive power of language to encapsulate complex decision-making processes. This transformation is made possible by leveraging recent breakthroughs in large language models (LLMs), which possess human-like ability to understand, process, and generate language-based information. Our initial experiments over tabular MDPs validate the effectiveness, efficiency, and interpretability of our NLRL framework.

% \textbf{Language Policy Evaluation (LPE)}: Instead of using scalar value to evaluate the state, LPE aims at learning language value function, which utilizes language description to assess the state. Language evaluation provides information-rich evaluation, including strategic analysis of environment feedback, richer semantic information, or comparison between different action choices, etc.

% \textbf{Language Policy Improvement (LPI)}: The lack of scalar value evaluation fails the traditional argmax operation. LPI conducts strategic language analysis of the relationship between language evaluation and the final task objective, choosing the most promising action to improve the policy.

\section{Preliminary of Reinforcement Learning}
Reinforcement Learning models the decision-making problem as a Markov Decision Process (MDP), defined by the state space $\mathcal{S}$, action space $\mathcal{A}$, probabilistic transition function $P: \mathcal{S} \times \mathcal{A} \times \mathcal{S} \rightarrow[0,1]$, discount factor $\gamma \in[0,1)$ and reward function $r: \mathcal{S} \times \mathcal{A} \rightarrow\left[-R_{\max }, R_{\max }\right]$. The goal of RL aims to learn a policy $\pi: \mathcal{S} \times \mathcal{A} \rightarrow[0,1]$, which measures the action $a$'s probability given the state $s$: $\pi(a|s)=\operatorname{Pr}\left(A_t=a \mid S_t=s\right)$. In decision-making task, the optimal policy tends to maximize the expected discounted cumulative reward: $\mathbb{E}_{\pi}\big[\sum_{t=0}^{\infty} \gamma^t r\left(\mathrm{s}_t, \mathrm{a}_t\right)\big]$. The state-action and state value functions are two key concepts that evaluate states and actions by RL objective's proxy:
$Q_\pi(s_t,a_t)=\mathbb{E}_{\mathrm{(s,a)}_{t+1: \infty} \sim P_\pi}\left[ \sum_{i=t}^\infty \gamma^{i-t} r\left(s_i,a_i\right) \mid s_t, a_t\right]$, $V_\pi(s_{t}) = \mathbb{E}_{\mathrm{(s,a)}_{t+1: \infty} \sim P_\pi}\left[\sum_{i=t}^\infty \gamma^{i-t} r\left(s_i,a_i\right) \mid s_t\right]$, where $P_{\pi}$ is the trajectory distribution given the policy $\pi$ and dynamic transition $P$. 

Given the definition of $V_\pi(s_{t})$, the relationship between temporally adjacent state's value can be derived as the Bellman expectation equation. Here is a one-step Bellman expectation equation:
\begin{equation}
\label{equ:one-step-bellman}
V_\pi(s_t)=E_{a_t\sim\pi_\theta}\Big[r(s_t,a_t)+\gamma E_{s_{t+1}\sim p(s_{t+1}|s_t,a_t)}[V_\pi(s_{t+1})]\Big], \forall s_{t} \in \mathcal{S}
\end{equation}

A similar equation can also be derived for $Q_\pi(s,a)$. Given these basic RL definitions and equations, we can illustrate how policy evaluation and policy improvement are conducted in GPI.

\textbf{Policy Evaluation}. The target of the policy evaluation process is to estimate state value function $V_\pi(s)$ or state-action value function $Q_\pi(s, a)$. For simplicity, we only utilize $V_\pi(s)$ in the following illustration. Two common value function estimation methods are the Monte Carlo (MC) estimate and the Temporal-Difference (TD) estimate \citep{sutton1988learning}. MC estimate leverages Monte-Carlo sampling over trajectories to construct unbiased estimation: $V_\pi(s_t)\approx \frac{1}{k}\sum_{n=1}^{K}[\sum_{i=t}^\infty \gamma^{i-t} r(s_i^n,a_i^n)]$. TD estimate relies on the temporal relationship shown in Equ.\ref{equ:one-step-bellman} to construct an estimation: $V_\pi(s_t)\approx \frac{1}{k}\sum_{n=1}^{K}[r(s_t,a_t^n)+\gamma V_\pi(s_{t+1}^n)]$, which can be seen as a bootstrap over next-state value function.

\textbf{Policy Improvement}. The policy improvement process aims to update and improve policy according to the result of policy evaluation. Specifically, it replaces the old policy $\pi_{\text{old}}$ with the new one $\pi_{\text{new}}$ to make the expected return increases: $V_{\pi_{\text {new }}}(s_0) \geq V_{\pi_{\text {old }}}(s_0)$. In the environment with small, discrete action spaces, such improvements can be achieved by greedily choosing the action that maximizes $Q_{\pi_{\text {old}}}(s, a)$ at each state: 
\begin{equation}
    \pi_{\text {new}}(\cdot \mid s)=\underset{\bar{\pi}(\cdot \mid s) \in \mathcal{P}(\mathcal{A})}{\arg \max } \mathbb{E}_{\mathrm{a} \sim \bar{\pi}}\left[Q_{\pi_{\text {old}}}(s, \mathrm{a})\right], \forall s \in \mathcal{S}
\end{equation}
Another improvement method involves applying policy gradient ascent \citep{sutton1999policy}. It parameterizes the policy $\pi_{\theta}$ with $\theta$. The analytical policy gradient can be derived as follows: 
\begin{equation}
\left.\nabla_\theta V\left(\pi_\theta\right)\right|_{\theta=\theta_{\text {old }}} = \mathbb{E}_{\mathrm{(s,a)} \sim P_{\pi_{\theta_{\text {old }}}}}\left[\left.\nabla_\theta\log{\pi_{\theta}(a|s)} Q_{\pi_{\theta_{\text {old }}}}(\mathrm{s}, \mathrm{a})\right|_{\theta=\theta_{\text {old }}}\right].
\end{equation}
By choosing a relatively small step-size $\alpha > 0$ to conduct gradient ascent: $\theta_{\text {new }}=\theta+\left.\alpha \nabla_\theta V_{\pi_{\theta}}(s_0)\right|_{\theta=\theta_{\text {old }}}$, we can guarantee the policy improvement: $V_{\pi_{\text {new }}}(s_0) \geq V_{\pi_{\text {old }}}(s_0)$.

\section{Natural Language Reinforcement Learning}
\label{sec:nlrl}
In contrast to the precise statistical models used in traditional RL, humans typically frame all elements—including task objectives, value evaluations, and strategic policies—within the form of natural language. This section aims to mimic how humans navigate decision-making tasks using natural language, aligning it with the concepts, definitions, and equations in traditional RL. Due to the inherent ambiguity of natural language, the equations presented here are not strictly derived from mathematical definitions. They are instead analogical and based on empirical insights of original RL concepts. We leave rigorous theory for future work.

\begin{figure}[t]
\vspace{-10pt}
    \centering
    \includegraphics[width=0.92\linewidth]{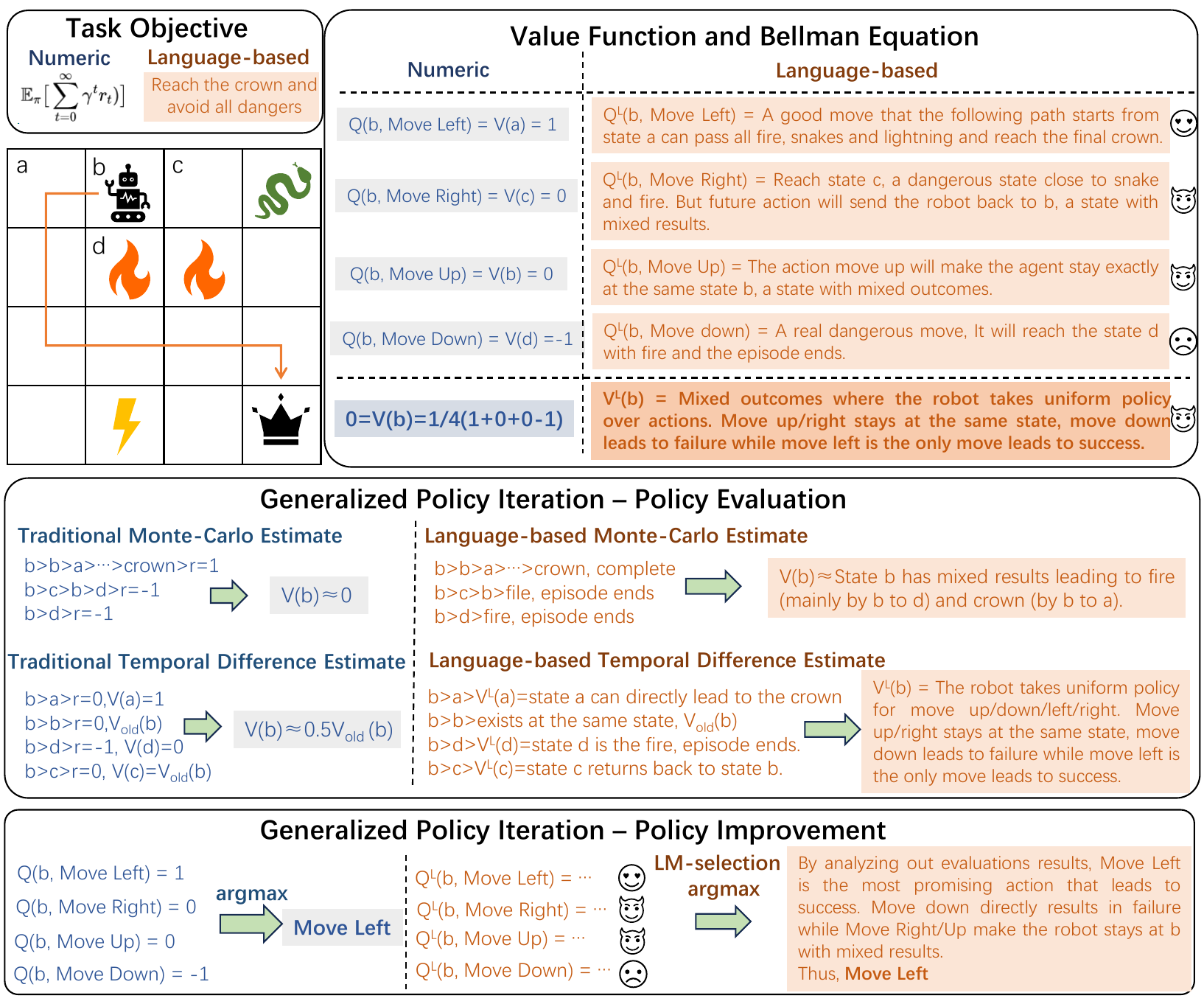}
    \vspace{-5pt}
    \caption{We present an illustrative example of grid-world MDP to show how NLRL and traditional RL differ for task objective, value function, Bellman equation, and generalized policy iteration. In this grid-world, the robot needs to reach the crown and avoid all dangers. We assume the robot policy takes optimal action at each non-terminal state, except a uniformly random policy at state b.}
    \label{fig:main_fig}
    \vspace{-10pt}
\end{figure}
\subsection{Definitions}
We start with definitions and equations in natural language RL to model human's behaviors. We provide Fig. \ref{fig:main_fig} with illustrative examples covering most concepts we will discuss in this section.

\textbf{Text-based MDP:} To conduct RL in natural language space, we first need to convert traditional MDP to text-based MDP, which leverages text and language descriptions to represent MDP's basic concepts, including state, action, and environment feedback (state transitions and reward).

\textbf{Language Task instruction:} Humans usually define a natural language task instruction $T_{L}$, like ``reaching the goal" or ``opening the door". Then, we denote a human metric by $F$ that measures the completeness of the task instruction given the trajectory description $D_{L}(\tau_{\pi})$, a language descriptor $D_{L}$ that can transform the trajectory distribution $\tau_{\pi}$ into its corresponding language description $D_L(\tau_{\pi})$. Based on the notation, the objective of NLRL is reformulated as 
\begin{equation}
    \max_{\pi} F(D_L(\tau_{\pi}), T_{L}),
\end{equation}

We are trying to optimize the policy so that the language description of the trajectory distribution $\tau_{\pi}$ can show high completeness of the task instruction.  

\textbf{Language Policy:} Instead of directly modeling action probability, humans determine the action by strategic thoughts, logical reasoning, and planning. Thus, we represent the policy on language as $\pi_{L}(a, c|s)$, which will first generate such thought process $c$, then the final action probability $\pi(a|s)$.

\textbf{Language Value Function:} Similar to the definition of $Q$ and $V$ in traditional RL, humans leverage language value function, relying on natural language evaluation to assess the policy effectiveness. The language state value function $V^{L}_\pi$ and language state-action value function $Q^{L}_\pi$ are defined as:
\begin{equation}
Q^{L}_\pi(s_t,a_t)=D\left(\mathrm{\left(s,a\right)}_{t+1: \infty}\sim P_\pi \mid s_t, a_t, T_{L}\right), V^{L}_\pi(s_{t}) = D\left(\mathrm{a_{t}, \left(s,a\right)}_{t+1: \infty}\sim P_\pi \mid s_t, T_{L}\right),
\end{equation}
Given the current state $s_t$ or state-action $(s_t,a_t)$, $Q^{L}_\pi$ and $V^{L}_\pi$ leverage language descriptions instead of scalar value to demonstrate the effectiveness of policy for achieving the task objective $T_{L}$. The language value functions are intuitively rich in the information of values and enhance interpretability rather than the traditional scalar-based value. It can represent the evaluation results from different perspectives, consisting of the underlying logic/thoughts, prediction/analysis of future outcomes, comparison among different actions, etc. 

\textbf{Language Bellman Equation:} In the Bellman equation, \textbf{the state evaluation value $V(s_{t})$, can be decomposed into two parts. Firstly the intermediate changes, which include immediate action $a_{t}$, reward $r_{t}$, and next state $s_{t+1}$. Secondly, the state evaluation $V(s_{t+1})$ over the next state.} Based on such intuition of decomposition, we introduce the language Bellman equation Equ.\ref{equ:langueg_bellman_exp} following the decomposition principle.
\begin{equation}
\label{equ:langueg_bellman_exp}
    V_\pi^{L}(s_t)=G_1^{a_{t},s_{t+1}\sim P_\pi}\Big(G_2\left(d\left(a_t,r\left(s_t,a_t\right),s_{t+1}),V_\pi^L(s_{t+1}\right)\right) \Big), \forall s_{t} \in \mathcal{S}
\end{equation}
where $d\left(a_t,r\left(s_t,a_t\right),s_{t+1})\right)$ depicts the language description of intermediate changes, $G_{1}$ and $G_{2}$ serves as two information aggregation function. Specifically, $G_{2}$ mimics the add `$+$' operation in the original Bellman equation, aggregating information from intermediate changes' descriptions and future evaluation given $a_{t}$ and $s_{t+1}$. $G_{1}$ takes the responsibility of the expectation operation $\mathbb{E}$, aggregating information from different $(a_{t},s_{t+1})$ pairs by sampling from trajectory distribution $P_{\pi}$.

\subsection{Language Generalized Policy Iteration}
Given definitions and equations, in this part, we introduce how language GPI is conducted. Refer to  Fig. \ref{fig:main_fig} with illustrative examples of language GPI.
\subsubsection{Language Policy Evaluation}
\label{sec:language_evaluation}
Language policy evaluation aims to estimate language value function $V^{L}_\pi$ and $Q^{L}_\pi$ for each state. We present how two classical estimations: MC and TD estimate work in language policy evaluation.

\textbf{Language Monte-Carlo Estimate.} Starting from the state $s_{t}$, MC estimate is conducted over text rollouts (i.e. $K$ full trajectories $\{a_{t}, \left(s,a\right)_{t+1: \infty}\}$) given the policy $\pi$. Since we cannot take the average operation in language space, we instead leverage language aggregator/descriptor $G_{1}$ to aggregate information over finite trajectories, approximating the expected evaluation.
\begin{equation}
\label{equ:mc_estimate}
    V^{L}_\pi(s_{t})\approx G_{1}\left(\left\{\mathrm{a_{t}^n, \left(s,a\right)}_{t+1: \infty}^{n}\right\}_{n=1}^{k}\right)
\end{equation}
\textbf{Language Temporal-Difference Estimate.}
TD estimate mainly relies on the one-step language Bellman equation illustrated in Equ. \ref{equ:langueg_bellman_exp}. Similar to MC estimate, we aggregate $K$ one-step samples to approximate the expected evaluation:
\begin{equation}
    V_\pi^{L}(s_t)\approx G_1\Big(\big\{G_2\big(d(s_t,a_t^n,r(s_t,a_t^n),s_{t+1}^n),V_\pi^L(s_{t+1}^n)\big)\big\}_{n=1}^K \Big), \forall s_{t} \in \mathcal{S}
\end{equation}

Language MC estimate is free from estimation ``bias" as it directly utilizes samples from complete trajectories. However, the MC method is prone to high ``variance" considering the significant variations in the long-term future steps. Such variability poses a challenge for the language aggregator $D$ in Equ.~\ref{equ:mc_estimate} to efficiently extract crucial information from diverse trajectories. On the contrary, while the inaccuracy of the next state evaluation $V_\pi^L(s_{t+1})$ can bring estimation ``bias" to TD estimate, they effectively reduce ``variance" by discarding future variations. $G_{1}$ and $G_{2}$ are only required to conduct simple one-step information aggregation with limited variations.\footnote{We use quotes for ``bias" and ``variance" to indicate that we draw on their conceptual essence, not their strict statistical definitions, to clarify concepts in NLRL.}

\subsubsection{Language Policy Improvement}
Similar to traditional policy improvement, the motivation of language policy improvement also aims to select actions that maximize the human task completeness function $F$:
\begin{equation}
\pi_{\text {new}}(\cdot \mid s)=\underset{\bar{\pi}(\cdot \mid s) \in \mathcal{P}(\mathcal{A})}{\arg \max } F(Q_{\pi_{\text{old}}}^{L}(s, a), T_{L}), \forall s \in \mathcal{S}
\end{equation}

As we mentioned, $F$ is typically a human measurement of task completeness, which is hard to quantify and take the argmax operation. Considering that $F$ largely depends on human prior knowledge, instead of mathematically optimizing it, we leverage the language analysis process $I$ to conduct policy optimization and select actions:
\begin{equation}
\pi_{\text {new}}(\cdot \mid s), c=I(Q_{\pi_{\text{old}}}^{L}(s, a), T_{L}), \bar{\pi}(\cdot \mid s) \in \mathcal{P}(\mathcal{A}), \forall s \in \mathcal{S}
\end{equation}
Language policy improvement conducts strategic analysis $C$ to generate the thought process $c$ and determine the most promising action as the new policy $\pi_{\text {new}}(\cdot \mid s)$. This analysis is mainly based on human's correlation judgment between the language evaluation $Q_{\pi_{\text{old}}}^{L}(s, a)$ and task objective $T_{L}$.

\subsection{Practical Implementation with large language models}
\label{practice}
Section \ref{sec:nlrl} demonstrates the philosophy of NLRL: transfer RL key concepts into its human natural language correspondence. To practically implement these key concepts, we require a model that can understand, process, and generate language information. The Large language model, trained with large-scale human language and knowledge corpus, can be a natural choice to help mimic human behaviors and implement these language RL components.

\textbf{LLMs as policy ($\pi_{L}$).} Many works adopted LLMs as the decision-making agent \citep{wang2023voyager, feng2023chessgpt, christianos2023pangu, yao2022react} with Chain-of-thought process \citep{wei2022cot0}. By setting proper instructions, LLMs can leverage natural language to describe their underlying thought for determining the action, akin to the human strategic thinking.

\textbf{LLMs as information extractor and aggregator ($G_{1}, G_{2}$) for concepts}. LLMs can be powerful information summarizers \citep{zhang2023benchmarking}, extractors \citep{xu2023large}, and aggregators to help us fuse intermediate changes and future language evaluations for language value function estimates. 

One core issue is to determine which kind of information we hope our LLMs to extract and aggregate. Inspired by works \citet{das2023state2explanation,sreedharan2020bridging, schut2023bridging, hayes2017improving} in the field of interpretable RL, we believe \textbf{Concept} can be the core. We adopt the illustration in \citet{das2023state2explanation} that concept is a general, task-objective oriented, and high-level abstraction grounded in human domain knowledge. For example, in the shortest path-finding problem, the path distance and available path sets are two concepts that are (1) high-level abstraction of the trajectories and are predefined in human prior knowledge, (2) generally applicable over different states, (3) directly relevant to the final task objective. Given these motivations, LLMs will try to aggregate and extract domain-specific concepts to form the value target information. Such concepts can be predefined by human's prior knowledge, or proposed by LLMs themselves.

\textbf{LLMs as value function approximator ($D_{L}, Q^{L},V^{L}$).} The key idea of value function approximation \citep{sutton1999policy} is to represent the value function with a parameterized
function instead of a table representation. Nowadays deep RL typically chooses neural networks that take the state as input and output one-dimension scalar value. For NLRL, the language value function approximation of $D_{L}, Q^{L},V^{L}$ can be naturally handled by (multi-modal) LLMs. LLMs can take in the features from the task's state, such as low-dimension statistics, text, and images, and output the corresponding language value judgment and descriptions. For the training of LLMs, we adopt the concept extractor/aggregator mentioned above to form MC or TD estimate (in Sec~\ref{sec:language_evaluation}), which can be used to finetune LLMs for better language critics. 

\textbf{LLMs as policy improvement operator ($I$).} With the chain-of-thought process and human prior knowledge, LLMs are better to determine the most promising action $\pi_{\text {new}}(\cdot \mid s)$ by taking language analysis $c$ over the correlation of language evaluation $Q_{\pi_{\text{old}}}^{L}(s, a)$ and task objective $T_{L}$. The underlying idea also aligns with some recent works \citep{kwon2023reward,rocamonde2023vision} that leverage LLMs or Vision-language models as the reward--they can accurately model the correlation.

\subsection{Discussions over other RL concepts}
To illustrate the versatility of the framework, we show several examples of how other fundamental RL concepts can be framed into NLRL.

\textbf{TD-$\lambda$} \citep{sutton1988learning}. Equ.~\ref{equ:langueg_bellman_exp} considers the one-step decomposition of the value function, or in the context of traditional RL, the TD(1) situation. A natural extension is to conduct an $n$-step expansion of the language value function and form the TD($n$) estimate. By further aggregating these different estimations (1 to $n$), we obtain the NLRL's TD-$\lambda$ estimation. TD-$\lambda$ under the language value function involves aggregating different language evaluations into the final language estimation.

\textbf{Exploration.} Exploration stands as a fundamental challenge in RL. Traditional RL approaches often utilize strategies that introduce randomnesses, such as the $\epsilon$-greedy method or entropy regularization \citep{sutton2018reinforcement}, to promote exploration. In contrast, NLRL offers a pathway for exploration that is guided by uncertainty. For instance, the language value function in NLRL is adept at incorporating models of state uncertainty. This allows for a deeper analytical process during the language policy improvement, where the strategy can be refined to reduce such uncertainties.
\begin{figure}[t]
    \centering
\includegraphics[width=\linewidth]{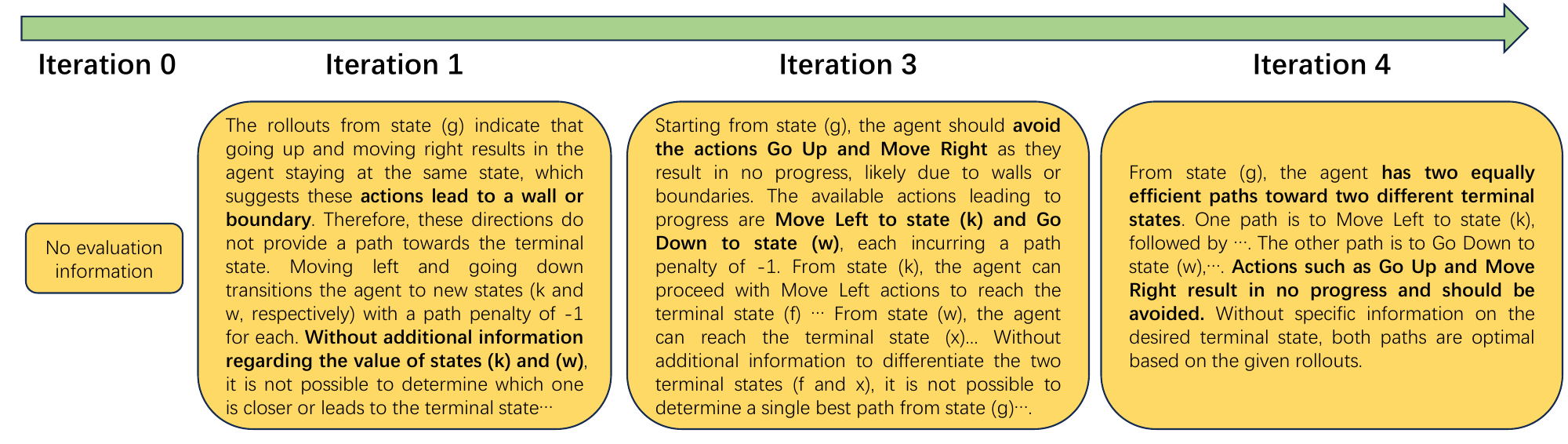}
\vspace{-20pt}
    \caption{How the language evaluation over state $g$ at (0,3) evolves across iterations. \textbf{Iter 0}: initial descriptions. \textbf{Iter 1}: Intermediate changes exclude action go up and move right. However, determining the optimal move is not possible without next-state evaluations. \textbf{Iter 3}: identify two optimal actions by information transmission from the goal. \textbf{Iter 4}: Evaluation over the state $g$ converges.}
    \label{fig:state_g_eval}
    \vspace{-10pt}
\end{figure}
\begin{figure}[t]
    \centering
\includegraphics[width=0.8\linewidth]{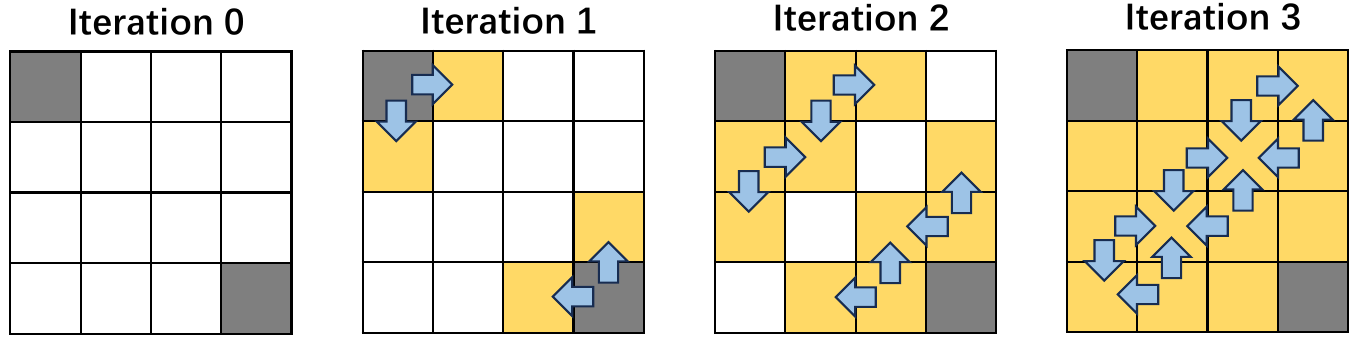}
    \caption{The information flows from the goal to every state across iterations. The yellow grid represents the grids where the goal information is transmitted. The blue array denotes the direction of information transmission by applying the one-step language Bellman update.}
    \label{fig:gridworld}
    \vspace{-10pt}
\end{figure}

\textbf{Episodic and Retrieval-based RL.} The concept of Episodic and Retrieval-based RL \citet{goyal2022retrieval,blundell2016model} is to keep explicit records of past trajectories, which will be used for reference to augment the decision-making ability in the new state. This aligns with the idea of Retrieval-augmented generation (RAG) \citep{lewis2020retrieval} in LLMs. By leveraging RAG over past trajectories, NLRL can also be extended to Retrieval-based RL.

\section{Experiments}
We conduct initial experiments over tabular MDP as proof-of-concept verifications. Refer to Appendix \ref{apx:exp} for more experimental results, settings and prompts. 
\label{headings}
\subsection{Warm-Up: Language Policy Evaluation in text GridWorld}
We start with the shortest path-finding problem, which is commonly adopted as an illustration for dynamic programming-based policy iteration algorithms. As shown in Fig. \ref{fig:gridworld} (Iteration 0), this is a $4\times4$ grid world and the goal is to reach two grey grids at (0,0) and (3,3) with the shortest path. We convert this environment to a text game, where we leverage natural language to represent the state, the transition, and the environment feedback such as path penalty and goal-reaching termination. We set the initial policy as a random policy that samples action from a uniform distribution over all available actions: $\pi(a|s)=\frac{1}{|\mathcal{A}|},\forall a \in\mathcal{A}, s \in\mathcal{S}$. We conduct language policy evaluation over this policy, similar to the policy evaluation process in the original policy iteration algorithm.

We initialize the language evaluation descriptions for all states within the language state value table as "No evaluation information", except terminal states, which are initialized with a predefined description specific to their terminal nature. Similar to the traditional policy evaluation process, each language policy evaluation iterates over all non-terminal states. For each state $s$ and action $a$, it leverages the one-step Bellman equation shown in Equ.~\ref{equ:langueg_bellman_exp} to calculate the new language state-action value function $Q^{L}(s,a)$ and $V^{L}(s)$. This evaluation is performed over four iterations to observe the evolution of the language state-action value function across iterations. We employ the \textit{GPT-4-preview-1106} model as the information aggregator for the language Bellman updates. Moreover, by converting coordinates to letters, we ensure that GPT-4 relies exclusively on environmental feedback for information gathering, thereby eliminating dependence on its inherent knowledge base.
\begin{table}[t]
\vspace{-18pt}
\small
    \centering
    \caption{Predefined concept names and corresponding definitions.}
    \label{tab:frozen_lake_concept}
    \begin{tabular}{cc}
        \toprule
         Concept name & Concept definition\\
         \midrule
         Important state & Record the important state positions such as goal position, and hole positions.\\
         Immediate risk & Identify immediate one-step failure. \\
        Future risk & Identify potential future failure paths (more than one step) or future bad states.\\
        Safest path & Generate the safest path starting from the given state and action.\\
        Final evaluation & The final evaluation of the policy in the given state.
        \\
        \bottomrule
    \end{tabular}
    \vspace{-10pt}
\end{table}

In Fig.~\ref{fig:state_g_eval}, we showcase the results of language evaluation for state $g$ at (0,3) across various iterations, serving as a representative example. For comprehensive results pertaining to different states, please see Appendix \ref{apx:exp}. Three key insights emerge from our analysis. First, despite that we prompt the GPT-4 model for policy evaluation, it inherently undertakes a policy improvement process, selecting the most advantageous action options. \textbf{This demonstrates GPT-4's capacity for language policy enhancement, as it interprets language evaluations and task goals.} 
Second, the evaluation outcomes in the final iteration correctly identify the optimal action for every state. \textbf{This confirms the language policy evaluation and iteration process's effectiveness and efficiency}, where merely four iterations of policy evaluation are sufficient to yield precise language evaluation and the respective optimal actions, significantly outperforming the conventional policy iteration method. 
Finally, the language evaluation outcomes elucidate the gradual transmission of goal-related information, with the example in Fig. \ref{fig:state_g_eval} illustrating how information from the goal flows to state $g$ over three evaluation iterations. Furthermore, Fig. \ref{fig:gridworld} is provided to depict the flow of information across iterations visually.

\begin{figure}[tbh]
\vspace{-10pt}
    \centering
    \includegraphics[width=0.98\linewidth]{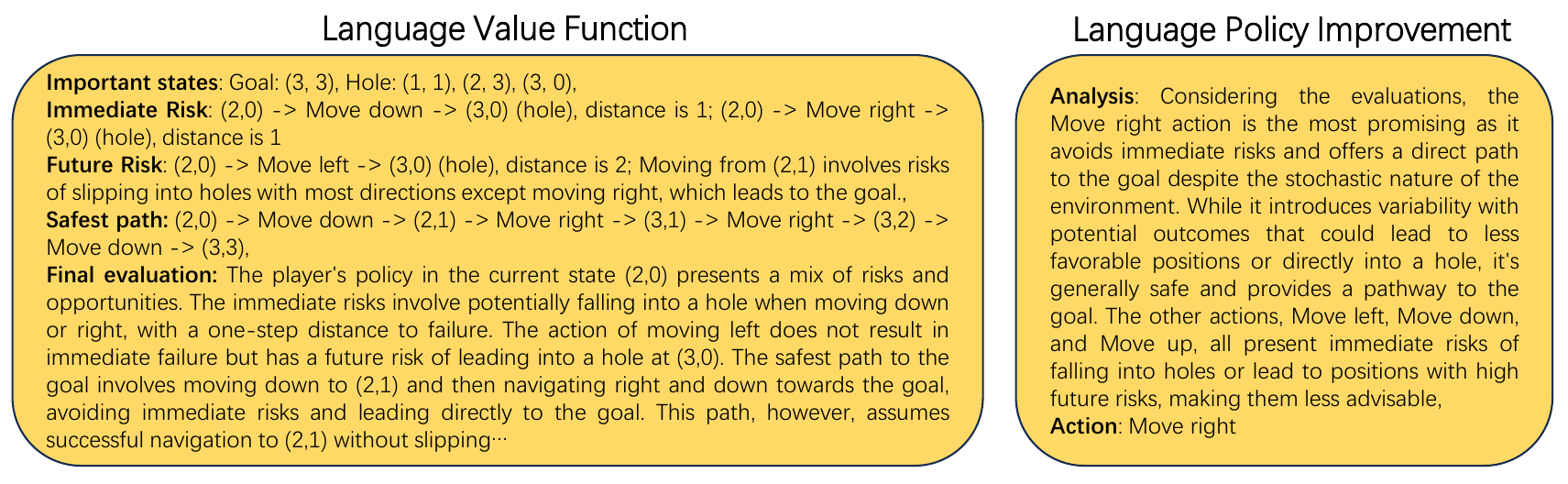}
    \vspace{-14pt}
    \caption{Frozen-Lake example of the language value function and language policy improvement. The language value function addresses the 5 predefined concepts, while the language policy improvements conduct chain-of-thought reasoning to determine the final action.}
    \label{fig:forzen_lake_example}
\end{figure}

\begin{table}[tbh]
\vspace{-18pt}
\small
    \centering
    \caption{Average state value over all states during the language policy iteration process, compared with the optimal policy's value on the right.}
    \label{tab:frozen_lake_result}
    \begin{tabular}{ccccccc}
        \toprule
         & Iter 0 & Iter 1 & Iter 2 & Iter 3 & Iter 4 & Optimal\\
         \midrule
         Average Value & 0.062 & 0.217 & 0.326 & 0.261 & 0.327 & 0.555
        \\
        \bottomrule
    \end{tabular}
    \vspace{-10pt}
\end{table}
\subsection{Language Policy evaluation and improvement in Stochastic Environment}
In our second experiment, we explore the Frozen-Lake environment from \citet{brockman2016openai}, a grid world designed to navigate an agent to a goal while avoiding lake holes, challenging due to the randomness in state transitions. Following the framework of \citet{sheng2023can}, we convert this environment into a text-based format. Our approach involves four iterations of language policy, each comprising an evaluation and an improvement phase.

Our initial experiments showed that the direct prompting method (used in our first experiment) easily failed the policy iteration process in the frozen-lake task. This is mainly because of the large number of trajectory variations, which can easily distract the LLM and disrupt the aggregation process. To tackle this problem, we leverage a more explicit method for aggregation: the concept-based language value function and information extraction discussed in Section \ref{practice}. We predefined several key concepts listed in Table \ref{tab:frozen_lake_concept} and prompted the \textit{GPT-4-0125-preview} model to generate evaluations and aggregate information based on these concepts.

Fig.~\ref{fig:forzen_lake_example} presents an example of the language value function and the language policy improvement process. The value function example clearly explains why the definition of task-relevant concepts can facilitate information aggregation. The GPT-4 can easily conduct explicit concept aggregation: it can update the path information by appending the newest move to the leftmost, or extract environment information by transmitting the coordinates of goals and holes.

In Table.~\ref{tab:frozen_lake_result} we provide the policy value for each iteration by leveraging frozen-lake's original reward function. The table presents an overall increasing trend, verifying the effectiveness of our language policy iteration process. Nevertheless, the system only attains $60\%$ efficiency compared to the optimal policy. Further analysis of the language value function pinpointed the issue: even the most advanced GPT-4 model is not immune to hallucination problems, occasionally generating inaccurate information. These inaccuracies can accumulate through successive Bellman updates, leading to incorrect evaluations and decisions. Efforts are underway to mitigate this instability through improved prompting strategies.

\section{Related work}
% In this section, we will contextualize our proposed NLRL framework amidst recent advancements in language model-based autonomous agents and interpretable reinforcement learning.

\textbf{Language Model Based Autonomous Agent.} 
% \textit{Large Language Models for Agent Planning}\\
Inspired by the strong emergent capabilities of LLMs in complex reasoning and planning scenarios \citep{gpt3,wei2022emergent,anil2023palm,OpenAI2023GPT4TR, feng2023alphazero}.
The field of language model-based autonomous agents have witnessed a growing trend of leveraging LLMs for high-level planning purposes. 
For text agents, ReAct\citep{yao2022react} leverages chain-of-thought planning about the intermediate goals and task-specific generation via few-shot prompting. Reflexion \citep{shinn2023reflexion} is built upon ReAct \citep{yao2022react} with self-reflexion, named as verbal reinforcement learning, to generate tips given the online feedback. This somehow aligns with the setting of language Monte-Carlo estimate where we extract core information/concepts by sampling multiple trajectories.
Another line of work like \citet{zhang2023benchmarking, xu2023large} leverages large language models to summarize and extract information to help fuse intermediate changes. 
In embodied agent research \citep{duan2022embodyaisurvey}, for instance, research for robot learning demonstrates large language models' potential in hierarchical planning and subgoal generation \citep{huang2022innermono, liang2023codeaspolicy}.

\textbf{Interpretable Reinforcement Learning.}
One of the major purposes of interpretable RL is to automatically seek explanations for non-AI experts. For instance, methods of concept-based explanations for sequential decision-making tasks. \citet{ji2023spatial} provide a concept-based explanation for 3D action recognition CovNets by clustering learned human interpretable features. \citet{sreedharan2020bridging} formulates concept-based explanations upon state preconditions and action costs, representing any factual statement a user associates with a given state. Similarly, \citet{hayes2017improving} uses logical formulas to summarize policies. 
Additionally, \citet{das2023state2explanation} trains a joint embedding model for state-action pairs and concept-based explanations.

\textbf{Learning from Language Feedback.} Our work is also related to learning from language feedback \citep{cheng2023llf}. \citet{cheng2023llf} focuses on how to benchmark algorithms while we aim to propose a new algorithm framework.
It has been investigated how to provide feedback to an RL agent's activity as part of the learning signal in addition to task rewards in the field of robotics and autonomous agents while in which most of the efforts were devoted to eliciting binary preference feedback\citep{biyik2018batch} or ranking-based labels from human\citep{sumers2021learning}. Our algorihtms can also benefit from different benchmarks \citep{liu2023agentbench,abdulhai2023lmrl,sheng2023can} that try to incorporate language feedback to develop RL policies. We are actively testing our algorihtms on these benchmarks.

\textbf{LLM as evaluation function.} Our language value function is also related to some work in the NLP field that leverages LLMs as evaluators or verifiers, including \citet{wang2023pandalm, li2023generative,jiang2023tigerscore, gao2024llm}. The main differences between our language value function with these works are two-fold. First, these works mainly focus on natural language tasks while we aim to tackle decision-making problems. Second, most of these works train their LLM evaluator by distilling strong models such as GPT-4\footnote{NLRL only leverages GPT-4 for information aggregation rather than GPT-4's evaluation ability.}, while we mainly rely on environmental feedback to generate synthetic evaluation data.
\section{Conclusion and Limitations}

In this work, we propose a comprehensive framework NLRL, which reformulates the RL process with natural language representation. We believe NLRL opens a new door for understanding and implementing RL algorithms in natural language space. In addition, NLRL also provides a potential way for generating high-quality language synthetic data, which could be crucial in training more advanced language models.

We are working on addressing the following limitations: (1) Currently, the LLM's hallucination issue severely influences the performance of NLRL. We will focus on addressing this problem to stabilize the NLRL's policy iteration process, (2) Our experiment is only limited to tabular MDP. We are working on scaling up our experiments beyond tabular settings, and (3) Currently, the performance of the language value function and the process of language policy iteration is measured purely by the policy's performance. We are working on a new metric to comprehensively evaluate the performance.

\bibliography{iclr2023_conference}
\bibliographystyle{iclr2023_conference}
\newpage
\appendix
\section{Experimental details}
\label{apx:exp}
\subsection{Prompts}
We present the prompts used in our shortest path-finding grid-world and Frozen-Lake environment.
\begin{mdframed}[style=mystyle, frametitle={The shortest path-finding prompts}]
\addcontentsline{toc}{paragraph}{The shortest path-finding prompts}
\small\ttfamily
\begin{spverbatim}
terminal_state_description = "The state is the terminal state. Your move ends here so you will not receive any negative path cost anymore in the future."

system_prompt = "You are a helpful assistant that strictly follows the user’s instructions.\n" \
"We are finding the shortest path towards the terminal state in gridworld. We use (letter) to represent different grids.\n" \
"To help us evaluate a given grid position in the shortest path finding task, we have an agent that can conduct actions to move to other grids. The agent has 4 actions: go up, move left, go down and move right. The agent conducts several one-step rollouts to explore such grid world.\n"
"For each one-step rollout starting from a given grid, it contains two sources of information, one about the descrption of intermediate change of action, state and reward, while the other one is about our evaluation of the new grid state.\n" \
"I need you to help me aggregate/summarize on these rollout information to concisely evaluate the given grid position."

rollout_prompt = "The current grid state is {current_state}. And here are several one-step rollout results starting from the current state {current_state}:\n\n"

trigger_prompt = "By aggregating the above rollout results, please concisely generate your evaluation of the current state {current_state}. Start your answer with the sentence 'Final evaluation of the current state:'."

action_description = f"You choose the {action_dict[action]} action. You receive a negative reward -1 as the path penalty. Now you are at a new state ({state_list[next_i][next_j]})."

\end{spverbatim}
\end{mdframed}

\begin{mdframed}[style=mystyle, frametitle={The prompts used in Frozen-Lake}]
\addcontentsline{toc}{paragraph}{The prompts used in Frozen-Lake}
\small\ttfamily
\begin{spverbatim}
### prompts for getting language state-action value function Q:

AGG_TD_STATE_INPUT_FORMAT_STR = """State: {state}
Action: {action}
Possible outcomes:
{agg_str}"""

AGG_TD_STATE_SYSTEM_PROMPT_FC = """You are a highly skilled assistant, committed to following user instructions precisely. Our task involves a player navigating a stochastic environment known as 'Frozen Lake,' with the ultimate goal of reaching the goal position while avoiding all holes. Your role is to assist in evaluating the player's action on the given state.
In this environment, the player makes decisions in each state, leading to potentially varied subsequent states due to the stochastic nature of the environment. We will provide you with a set of possible outcomes resulting from one action.
For each outcome, you will receive two key pieces of information:

1. A detailed description of the immediate changes in state and the accompanying reward. This narrative will offer insights into the direct consequences of the player's actions.
2. An assessment of the player's policy at the newly reached state. This evaluation will include a future analysis of potential outcomes stemming from the new state.

### Your task
Your task is to synthesize these elements - the future state evaluations and the intermediate state descriptions - to construct a comprehensive evaluation of the player's action given current state.

### Json keys
You need to generate the analysis in the json format, here are the keys:
    1. Important states: Record the important state positions such as goal position, hole positions or other important states.
    2. Immediate Risk: Identify one-step failure (the action can directly result in failure). Generate None if no paths found. Here is an example if you are evaluating over state (4,1): (4,1) -> Move left -> (4,0) (hole), distance is 1.
    3. Future Risk: Identify potential future failure paths (more than one step) or future bad states. Generate None if no paths found.
    4. Safest: Generate the safest path starting from the given state and action. Output None if there are no safe paths. Here is an example if you are evaluating over state (4,1): (4,1) -> Move right -> (4,2) -> Move down -> (6,2), this is the goal.
    5. Final evaluation: Based on the above infomation, generate your final evaluation of the policy in the given state.

Make your analysis and evaluation concise and short.
One criteria is that a good action should not result in immediate risk.
"""

### prompts for getting language state value function V:

AGG_ACTION_INPUT_FORMAT_STR = """State: {state}
Possible actions and evaluations:

{agg_str}"""

AGG_ACTION_SYSTEM_PROMPT_FC = """Our task involves a player navigating a stochastic environment known as 'Frozen Lake,' with the ultimate goal of reaching an unknown destination. Your role is to assist in evaluating the player's action.
In this environment, the player makes decisions in each state, leading to potentially varied subsequent states due to the stochastic nature of the environment.
You are tasked with evaluating a player's stochastic policy in 'Frozen Lake'. The player has multiple actions to choose from in each state.
You will be provided with current state infromation and action choices with their corresponding action evaluations (in json format).

### Your task
Your task is to synthesize the effectiveness of different actions and combine these insights to assess the overall effectiveness of the player's policy in the current state.

### Json keys
You need to generate the analysis in the json format, here are the keys:
    1. Important states: Record the important state positions such as goal position, hole positions or other important states.
    2. Immediate Risk: Identify one-step failure (the action can directly result in failure). Here is an example if you are evaluating over state (4,1): (4,1) -> Move left -> (4,0) (hole), distance is 1.
    3. Future Risk: Identify potential future failure paths (more than one step) or future bad states.
    4. Safest path: Generate the safest path starting from the given state. Output None if there are no safe paths. Here is an example: (4,1) -> Move right -> (4,2) -> Move down -> (6,2), this is the goal.
    5. Final evaluation: Based on the above infomation, generate your final evaluation of the policy in the given state.

Make your analysis and evaluation concise and short.
One criteria is that a good action should not result in immediate failure and hole.
"""
\end{spverbatim}
\end{mdframed}

\subsection{Experimental settings and results}
For the shortest path-finding problem, we leverage \textit{GPT-4-preview-1106}, with temperature=1.0. Here we present the full language evaluation results for the whole 4 iterations. Each iteration has one $4\times4$ matrix for language state value function.
\begin{mdframed}[style=mystyle, frametitle={Language evaluation results in grid-world across iterations}]
\addcontentsline{toc}{paragraph}{Language evaluation results in grid-world across iterations}
\small\ttfamily
\begin{spverbatim}
### Iteration 1

[['The state is the terminal state. Your move ends here so you will not receive any negative path cost anymore in the future.',
  'The best action to take from state (b) is the Move Left action leading to state (f), which is the terminal state. This outcome is based on the provided rollouts, as it immediately leads to the terminal state without incurring any additional negative path costs.',
  'State (k) has four possible actions, each resulting in a -1 penalty. Going up keeps the agent in the same state (k), going left takes the agent to state (b), going down takes the agent to state (m), and moving right takes the agent to state (g). Without additional evaluation information for states (b), (m), and (g), it is not possible to determine the most beneficial action from state (k) towards the terminal state.',
  'The rollouts from state (g) indicate that going up and moving right results in the agent staying at the same state, which suggests these actions lead to a wall or boundary. Therefore, these directions do not provide a path towards the terminal state. Moving left and going down transitions the agent to new states (k and w, respectively) with a path penalty of -1 for each. Without additional information regarding the value of states (k) and (w), it is not possible to determine which one is closer or leads to the terminal state. However, since these are the only actions that lead to new states, further exploration from state (k) or state (w) would be necessary to find the shortest path toward the terminal state.'],
 ['The current state (z) has four possible actions. The Go Up action leads directly to the terminal state (f) with a path cost of -1, which is the optimal move since it results in reaching the terminal state immediately. The Move Left action results in no change, staying at (z) with a path cost of -1. Both Go Down and Move Right actions lead to new states (r) and (t) respectively with a path cost of -1; however, their values or positions in relation to the terminal state are unknown from the given rollouts. Therefore, given the available information, the best action from state (z) is to Go Up to reach the terminal state (f) immediately.',
  'With the information provided, it is not possible to determine the optimal direction for the shortest path from state (t) as there are no evaluations for the resulting states (b), (z), (y), and (m) after performing the actions Go Up, Move Left, Go Down, and Move Right respectively. All actions from state (t) incur a path penalty of -1, but without further information about the desirability or utility of subsequent states, we cannot conclude which action would be part of the shortest path to the terminal state.',
  'Since all one-step rollouts from the current state (m) result in a negative reward of -1 and there is no further evaluation information available for the new states (k), (t), (q), and (w), we cannot determine which direction leads to the shortest path towards the terminal state. Further information about the states following (k), (t), (q), and (w) is required to make an informed evaluation.',
  'The one-step rollouts from state (w) lead to states (g), (m), (n), and back to (w) with a uniform negative reward of -1 for the path penalty. However, there is no further evaluation information provided for the new states (g), (m), or (n). Therefore, we cannot determine which of these one-step transitions, if any, leads closer to the terminal state. Additional evaluation information would be needed to determine the most promising action from state (w).'],
 ["Considering the one-step rollouts from state (r) and without any further evaluation information for states (z), (j), and (y), all actions from state (r) result in a negative reward of -1 due to the path penalty. There is currently no information to distinguish the actions based on the follow-up states' values. More information or further exploration would be needed to determine the optimal action from state (r).",
  'Final evaluation of the current state (y): With the available information from the one-step rollouts, it is not possible to determine which action leads to the shortest path towards the terminal state. Each action taken from the current state (y) incurs a path penalty of -1, and there is no further evaluation information given for the resulting new states (t), (r), (d), and (q). Therefore, additional information or exploration is needed to ascertain the most beneficial action from state (y).',
  'Since all actions from state (q) lead to different new states (m, y, e, n) with a uniform negative reward of -1 and no further evaluation information provided, the current state (q) cannot be distinctly evaluated over the other states in terms of guiding to a terminal state or providing a better reward path. Without additional information or context regarding the final goal state or additional state rewards, all actions from the current state appear to be of equal value based on the given one-step rollouts.',
  'The best action to take from state (n) is to Go Down, moving to state (x), as it leads to the terminal state, thereby ending the path and avoiding any further negative path cost.'],
 ['Considering the outcomes of the rollouts from state (j), we have four attempted actions with associated penalties but no further evaluative information about the resulting states. Rollouts 1 and 2 return the agent to the same state (j), so these actions do not progress towards the terminal state. Rollouts 0 and 3 move the agent to new states (r) and (d) respectively, both with a path penalty of -1. Without additional information regarding states (r) and (d), we cannot determine which, if either, leads closer to the terminal state. Based on current data, state (j) is a decision point with yet unspecified potential, requiring further evaluation of subsequent states to ascertain the optimal action.',
  'Without any evaluation information on the resulting states (y), (j), (d), and (e) after the one-step rollouts from state (d), it is not possible to determine a preferred action or the value of state (d). Each action incurred a negative reward of -1, and further information is required to assess which action leads us closer to the terminal state.',
  'Based on the provided one-step rollouts, the most favorable action from state (e) is Move Right, leading directly to the terminal state (x) with a path cost of -1 and no further negative costs incurred. Therefore, the path through state (x) is the shortest among the explored options.',
  'The state is the terminal state. Your move ends here so you will not receive any negative path cost anymore in the future.']]

### Iteration 2

[['The state is the terminal state. Your move ends here so you will not receive any negative path cost anymore in the future.',
  'The optimal action from state (b) is the Move Left action as it immediately leads to the terminal state (f), ensuring the least cost path with no further negative path penalties.',
  'The most promising action from state (k) appears to be the Move Left action, which takes the agent to state (b). From state (b), the agent can then take the Move Left action again to reach state (f), which is the terminal state. This sequence of actions leads to the terminal state with a total penalty of -2, based on the given rollout information. Other actions from state (k) either keep the agent in the same state or take the agent to states without a clear path to the terminal state, requiring further exploration for a definitive evaluation.',
  'Actions going up and moving right from state (g) do not provide progress towards the terminal state as the agent remains in state (g), indicating these directions likely lead to walls or boundaries. Actions moving left and going down lead to new states (k and w respectively) with a path penalty of -1 for each. Since there are no further details provided to compare the potential benefits of states (k) and (w), it is not possible to determine which direction is more favorable from the current information. Both state (k) and state (w) would need further exploration to ascertain the most effective route towards the terminal state.'],
 ['The best action from state (z) is to Go Up to reach the terminal state (f) immediately with a path cost of -1, making it the optimal move from the current state given the rollout information provided.',
  'Final evaluation of the current state (t): The best action to take from state (t), based on the provided one-step rollouts, is the Move Left action, leading directly to state (z). From state (z), the optimal follow-up action is the Go Up action, which takes the agent to the terminal state (f) with the lowest cumulative path cost from the current state. This path incurs a total negative reward of -2 (one for moving left from (t) to (z), and one for going up from (z) to (f)), and leads to the terminal state with fewer steps compared to other available actions from state (t).',
  "Without additional evaluative information or a known terminal state, it is not possible to determine a preferred action from state (m), as all actions leading to states (k), (t), (q), and (w) result in a negative reward of -1 and each of these states has subsequent actions that also produce a negative reward of -1 without further direction. Additional information about the desirability of each state or knowledge of the terminal state's location is required to make a valuable assessment for the shortest path strategy.",
  'Based on the available one-step rollouts from state (w), the best action appears to be Go Down, leading to state (n). This state offers a pathway to state (x) without any further negative path costs, as state (x) is the terminal state. Actions such as Go Up, Move Left, and Move Right either lead to states with uncertain paths to the terminal or back to the same state (w), resulting in additional negative rewards without progression towards the goal. Therefore, the optimal move from state (w) is to Go Down to state (n), followed by another Go Down to reach the terminal state (x).'],
 ['Final evaluation of the current state (r): The best action to take from state (r) based on the given rollouts is to Go Up, which leads to state (z). From state (z), the agent can take the Go Up action to immediately reach the terminal state (f) with an overall path cost of -2, which is the minimum known cost for reaching the terminal state from (r).',
  "Final evaluation of the current state (y): Based on the one-step rollouts provided for state (y), all adjacent states (t, r, d, q) are reachable with a uniform negative reward of -1, which suggests that merely reaching these states does not provide sufficient information to distinguish the optimal move for the shortest path. The evaluation of the current grid position (y) requires additional information about the terminal state's location, the grid's wider layout, and potential rewards or penalties beyond the immediate adjacent states. As such, without further data or a defined terminal state, it is not possible to conclusively determine the value of state (y) or recommend the best action to progress towards the terminal state.",
  "Final evaluation of the current state (q): The most beneficial action to take from state (q) is to conduct the action 'Go Down' as per Rollout 2, since this action leads to state (e) and from there the agent can move right to the terminal state (x) with an overall path cost of -2, which appears to be the shortest path based on the given information.",
  'The most promising action from state (n) is to Go Down, as it directly leads to the terminal state (x), which stops accruing any further negative path cost. Other actions such as Go Up, Move Left, and Move Right lead to states that either have no additional evaluative information to suggest they are closer to the terminal state or in the case of Move Right, lead back to the starting state (n). Hence, the optimal move from (n) is the Go Down action leading to the terminal state (x).'],
 ['Final evaluation of the current state (j): Given the available data from the one-step rollouts, it appears the agent remains at state (j) when selecting either the Move Left action or the Go Down action, both resulting in a negative reward of -1 without progress toward the terminal state. The Go Up action leads to state (r), and the Move Right action leads to state (d), with both actions also incurring a negative reward of -1. However, there is insufficient data to determine whether state (r) or state (d) brings the agent closer to the terminal state. Consequently, more information and exploration are required to discern the optimal action from state (j).',
  'Based on the one-step rollout information provided, state (d) presents as a decision node in the grid world with uncertain outcomes. However, given that one of the one-step rollouts from state (d) leads to state (e), which in turn has a known path to the terminal state (x), we can deduce that moving right to state (e) and then once more to the terminal state (x) is currently the shortest path identifiable from state (d). Thus, the best current action from state (d) is to move right based on the limited information available. Further exploration from states (y), (j), and from state (d) itself may need to be conducted to conclusively determine the overall shortest path to the terminal state within the entire grid world.',
  'Based on one-step rollouts from state (e), the Move Right action leading directly to the terminal state (x) is the most favorable action, with a path penality of -1. Moving up to state (q), left to state (d), or down to remain in state (e) incurs a negative reward of -1 without leading to the terminal state, making these actions less desirable in the context of reaching the terminal state as quickly as possible.',
  'The state is the terminal state. Your move ends here so you will not receive any negative path cost anymore in the future.']]

### Iteration 3

[['The state is the terminal state. Your move ends here so you will not receive any negative path cost anymore in the future.',
  'Final evaluation of the current state (b): The optimal action from state (b) is the Move Left action as it leads directly to the terminal state (f) with a single path penalty of -1, which is the least cost path with no further negative path penalties according to the provided one-step rollouts.',
  'The most promising action from state (k) is the Move Left action, leading to state (b), which offers a clear path to the terminal state (f) with a total penalty of -2. Other actions such as Go Up, Go Down, and Move Right lead to states (k), (m), and (g) respectively, and do not present a clear or direct path to the terminal state based on the given rollout information.',
  'Starting from state (g), the agent should avoid the actions Go Up and Move Right as they result in no progress, likely due to walls or boundaries. The available actions leading to progress are Move Left to state (k) and Go Down to state (w), each incurring a path penalty of -1. From state (k), the agent can proceed with Move Left actions to reach the terminal state (f) with an additional -1 penalty, totaling a penalty of -2 from the current state (g). From state (w), the agent can reach the terminal state (x) by performing the Go Down action twice without any further penalties, also totaling a penalty of -2 from (g). Without additional information to differentiate the two terminal states (f and x), it is not possible to determine a single best path from state (g). Both paths to (f) and (x) seem equally efficient based on the given rollouts. Further rollouts or information about the specific terminal state goal would be necessary to make a conclusive decision on the optimal path.'],
 ['Final evaluation of the current state (z): The best action to take from state (z) is the Go Up action, which leads directly to the terminal state (f) with an immediate reward of -1 for the path cost. This is the optimal move as it minimizes the path cost and reaches the terminal state in the fewest number of steps according to the given rollouts.',
  'Final evaluation of the current state (t): Given the results of the one-step rollouts, the Go Up action leading to state (b) appears to be the most promising, as the subsequent action from there is Move Left, which directly leads to the terminal state (f) with no further penalties. This suggests that the optimal path from the current state (t) involves the Go Up action followed by Move Left, minimizing the path cost to reach the terminal state efficiently. Rollouts from other actions did not provide a clear or immediate path to the terminal state.',
  'Final evaluation of the current state (m): Considering the one-step rollouts, the Go Up action that leads to state (k) appears to be the most promising initial action. From state (k), the agent can then Move Left to state (b) and again Move Left to reach the terminal state (f). This sequence of moves results in a total path penalty of -3, offering a clear and direct path to the terminal state, which based on the given rollout information is the shortest path with the fewest steps from the current state (m).',
  'The optimal move from state (w) is to Go Down, transitioning the agent to state (n), which allows for a subsequent Go Down action to reach the terminal state (x) without incurring further negative path penalties. This is the shortest path to the terminal state based on the provided rollout information.'],
 ['Final evaluation of the current state (r): Based on the available one-step rollouts from the current state (r), it is determined that the optimal action is to Go Up, leading to state (z). From state (z), the agent can Go Up again to reach the terminal state (f) with a total path cost of -2, which is the shortest known path to the terminal state from (r). Other actions lead to states (r), (j), and (y), which have not been determined to provide a more efficient path to the terminal state within the scope of the information provided. Therefore, the recommended action from state (r) is to Go Up.',
  'Final evaluation of the current state (y): Based on the one-step rollouts conducted from state (y), the optimal action to take is the Move Left action to reach state (r). From state (r), the best subsequent action is to Go Up, arriving at state (z). Finally, from state (z), the Go Up action will lead to the terminal state (f) with the lowest cumulative path cost of -2, implying that this route is the shortest path to the terminal state from the current state (y).',
  'Final evaluation of the current state (q): Based on the provided one-step rollouts, it appears that the shortest path toward the terminal state can be achieved by first selecting the Go Down action to transition to state (e), and subsequently the Move Right action to reach the terminal state (x). The other actions evaluated during the rollouts did not provide a path to the terminal state and consistently resulted in a negative reward of -1 without additional strategic information to justify their selection. Therefore, the path from state (q) to state (e), and then to the terminal state (x) is the most advantageous sequence of actions based on the available data.',
  'Final evaluation of the current state (n): The best action for the agent to take from the current state (n) is to Go Down, according to Rollout 2. This action leads directly to the terminal state (x) with a path penalty of -1, which is the shortest path with the least negative reward based on the available rollouts.'],
 ['Final evaluation of the current state (j): Based on the available one-step rollout data, the agent remains at state (j) when selecting either the Move Left action or the Go Down action, both incurring a path penalty of -1 without advancement toward the terminal state. The choices of Go Up or Move Right from state (j) also result in a path penalty of -1, but lead to states (r) and (d) respectively. Rollout data suggests that state (r) is two steps away from the terminal state (f) with a known path cost of -2. State (d) is connected to a path that leads to the terminal state (x), but the exact path cost from state (d) to the terminal state (x) remains unclear. With the data at hand, and considering the lower known cost to reach the terminal state from state (r), the Go Up action from state (j) to state (r), and subsequently to state (z) and then to the terminal state (f), appears to be the most promising path. Additional exploration is needed to confirm the optimal route fully, as well as to ascertain the cost and route details for reaching the terminal state (x) from state (d).',
  'Final evaluation of the current state (d): Based on the gathered data from the one-step rollouts, the shortest confirmed path from state (d) towards the terminal state involves performing the Move Right action to transition to state (e), and then conducting another Move Right action to reach the terminal state (x), with each action incurring a negative reward of -1 as a path penalty. Any further actions from state (d) have not been concretely linked to a path towards the terminal state, necessitating additional exploration for a comprehensive shortest path analysis. However, with the current information, moving right twice from state (d) to reach state (x) appears to be the most optimal known path to the terminal state.',
  'Final evaluation of the current state (e): The most beneficial action to take from state (e) is to conduct the Move Right action which leads directly to the terminal state (x) with a path penalty of -1. This is the shortest path to the terminal state based on the one-step rollout results provided.',
  'The state is the terminal state. Your move ends here so you will not receive any negative path cost anymore in the future.']]

### Iteration 4

[['The state is the terminal state. Your move ends here so you will not receive any negative path cost anymore in the future.',
  'The optimal action from state (b) is the Move Left action as it leads directly to the terminal state (f) with a single path penalty of -1, which is the best option according to the current one-step rollouts provided.',
  'Final evaluation of the current state (k): The optimal action from state (k) is the Move Left action, resulting in a move to state (b) and then following another Move Left to reach the terminal state (f). This path incurs a total penalty of -2, which is the least negative penalty according to the provided one-step rollouts, offering the clearest and most direct path to the terminal state from the current state (k).',
  'From state (g), the agent has two equally efficient paths toward two different terminal states. One path is to Move Left to state (k), followed by another Move Left to the terminal state (f), totaling a path penalty of -2. The other path is to Go Down to state (w), followed by two Go Down actions reaching the terminal state (x), also with a total path penalty of -2. Actions such as Go Up and Move Right result in no progress and should be avoided. Without specific information on the desired terminal state, both paths are optimal based on the given rollouts.'],
 ['Final evaluation of the current state (z): The best action to take from state (z) is the Go Up action, which directly leads to the terminal state (f) with an immediate reward of -1 for the path cost. This is the optimal move as it reaches the terminal state in the fewest number of steps with the least negative reward based on the given rollouts.',
  'Final evaluation of the current state (t): Given the outcomes of the one-step rollouts, the optimal action to take from state (t) is the Move Left action, resulting in the agent moving to state (z). From state (z), the agent can then Go Up to directly reach the terminal state (f). This sequence of actions incurs a total path penalty of -2, which is the lowest cumulative cost among all explored options, thus providing the shortest path to the terminal state from the current state (t).',
  'Final evaluation of the current state (m): After examining the one-step rollout outcomes, the most promising action from state (m) seems to be the Go Up action, leading to state (k). From state (k), the subsequent Move Left action would take the agent to state (b), which presents a direct and clear path to the terminal state (f), incurring a total path penalty of -2. This path is suggested due to the strategic information given in Rollout 0 that indicates a clear path to the terminal state from state (b). Other actions such as Move Left, Go Down, and Move Right lead to states (t), (q), and (w) respectively, but none of these rollouts suggest a better path to the terminal state than the one starting with the Go Up action.',
  'Final evaluation of the current state (w): The optimal action for an agent at state (w) appears to be the Go Down action, resulting in reaching state (n) with a -1 path penalty. Then, from state (n), the agent can perform another Go Down action, leading directly to the terminal state (x) with no additional path penalties. This sequence of actions results in a total path penalty of -2 and offers the shortest path to a terminal state from the current state (w) based on the given rollouts.'],
 ['Final evaluation of the current state (r): Based on the one-step rollouts conducted, the most efficient action from state (r) is to Go Up to state (z). From state (z), the Go Up action again will lead directly to the terminal state (f), with a total path cost of -2. This route offers the optimal path to the terminal state considering the rollout information provided.',
  'Final evaluation of the current state (y): Considering the information available from the one-step rollouts, the state with the most promising subsequent actions leading directly to the terminal state is state (t), which is reached by the Go Up action. From state (t), the agent can Move Left to directly reach the terminal state (f). Therefore, the recommended action from state (y) is to Go Up to state (t) and then Move Left towards the terminal state (f) for the shortest known path given the current data.',
  'Final evaluation of the current state (q): According to the rollout results, the most promising move from (q) appears to be the Go Down action leading to state (e), as it results in reaching the terminal state (x) with the lowest cumulative negative reward of -1, indicating the shortest path from the current state. Therefore, the next action should be Go Down.',
  'Final evaluation of the current state (n): The optimal action for the agent from state (n) is to Go Down, as demonstrated by Rollout 2. This action transitions the agent directly to the terminal state (x) with a path penalty of -1, representing the shortest path with the least negative reward. None of the other actions evaluated in the rollouts provide a more advantageous route to the terminal state from the current state (n).'],
 ['Final evaluation of the current state (j): The optimal move from state (j) based on the available data is to Go Up to state (r). From state (r), the agent can Go Up again to reach state (z) and subsequently move up to the terminal state (f) with a total known path cost of -3. Other actions such as Move Left and Go Down lead back to state (j) with a path penalty of -1, and do not progress towards the terminal state. The Move Right action from state (j) leads to state (d), which connects to terminal state (x) via two Move Right actions with a total path cost of -2. However, since the total known path cost to reach terminal state (f) from state (j) is lower, the route from state (j) to state (r) to state (z) and finally to terminal state (f) is currently the most promising path for the shortest route based on the given data.',
  'Based on the one-step rollouts provided, the optimal action from state (d) is to Move Right, transitioning to state (e), and then to perform another Move Right action to reach the terminal state (x). This path incurs a total path penalty of -2, representing the shortest confirmed route to a terminal state from the current state (d).',
  'Final evaluation of the current state (e): Based on the one-step rollout results, the Move Right action from state (e) immediately leads to the terminal state (x) with a path penalty of -1. This is the optimal action to take from state (e) as it directly achieves the goal with the minimal penalty, rendering it the shortest path to the terminal state according to the data from the rollouts.',
  'The state is the terminal state. Your move ends here so you will not receive any negative path cost anymore in the future.']]
\end{spverbatim}
\end{mdframed}
\newpage
For the frozen-lake experiment, we leverage \textit{GPT-4-preview-0125} with temperature as 0. In Fig. \ref{apx:fig:frozen-lake} we present the policy value for each state across different iterations. For the language evaluation data, we will open-source them on Huggingface.
\begin{figure}
    \centering
    \includegraphics[width=\linewidth]{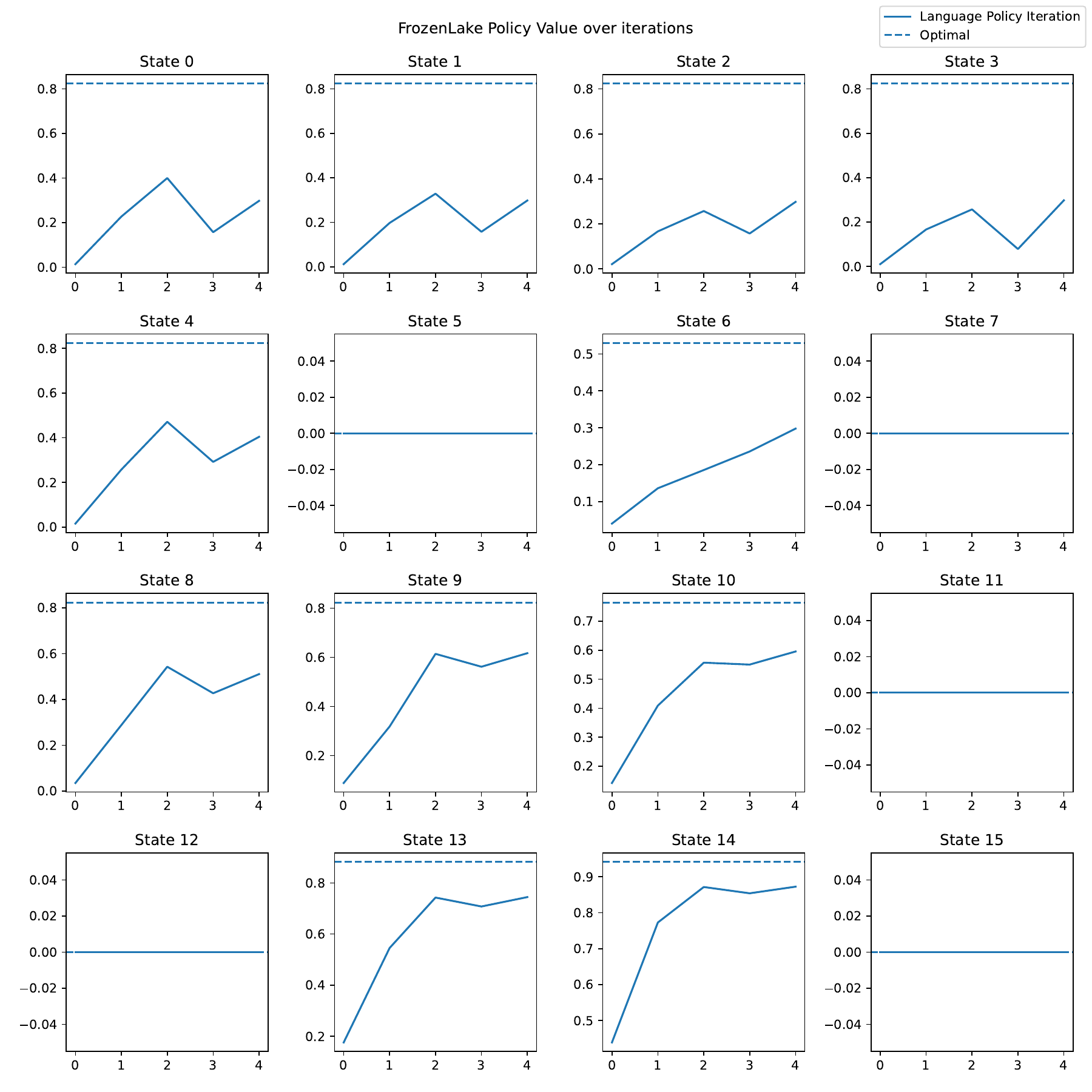}
    \caption{The policy value at each state in Frozen-Lake.}
    \label{apx:fig:frozen-lake}
\end{figure}

\end{document}

%% file: math_commands.tex
\usepackage{amsmath,amsfonts,bm}

\def\eqref#1{equation~\ref{#1}}
\def\1{\bm{1}}

\DeclareMathAlphabet{\mathsfit}{\encodingdefault}{\sfdefault}{m}{sl}
\SetMathAlphabet{\mathsfit}{bold}{\encodingdefault}{\sfdefault}{bx}{n}